# Dynamic Fault Analysis in Substations Based on Knowledge Graphs

Weiwei Li[1], Xing Liu[2], Wei Wang[2], Lu Chen[2], Sizhe Li[1], Hui Fan[1]
1. Nanjing University of Finance & Economics, Nanjing, Jiangsu, China, 210000
2. Suqian Power Supply Company, Suqian, Jiangsu, China, 223800

**Abstract:** To address the challenge of identifying hidden danger in substations from unstructured text, a novel dynamic analysis method is proposed. We first extract relevant information from the unstructured text, and then leverages a flexible distributed search engine built on Elastic-Search to handle the data. Following this, the hidden Markov model is employed to train the data within the engine. The Viterbi algorithm is integrated to decipher the hidden state sequences, facilitating the segmentation and labeling of entities related to hidden dangers. The final step involves using the Neo4j graph database to dynamically create a knowledge graph that visualizes hidden dangers in the substation. The effectiveness of the proposed method is demonstrated through a case analysis from a specific substation with hidden dangers revealed in the text records.

In recent years, the safety work of electric power enterprises has been steadily promoted, the construction of safety culture has been continuously deepened, and the level of safety management has been continuously upgraded[1-2]. However, under the high-pressure situation of safety control, the hidden dangers of substation equipment and the index of hidden dangers are still high. Because of the omnipresence of potential safety hazards in the production process of substations, involving personnel hazards, equipment and facilities hazards, fire hazards, electric power safety hazards, etc., it is necessary for operation and maintenance management personnel to discover and standardize the protection in time, so as to reduce the probability of safety accidents occurring[3-5]. In practice, substation hidden danger information is recorded by manually entering the hidden danger investigation and management form, and the unstructured nature of the hidden danger records in substations makes it very difficult to analyze the hidden danger. Although there are relevant electric power specifications that summarize the components and corresponding phenomena that may cause hidden problems in the form of tables, the complexity and diversity of hidden problems make it difficult to summarize them comprehensively in the tables in the specifications[6].

A lot of research has been done on text processing in the field of power system. Reference [7-8] established a semantic framework based on artificial experience, and filled in the semantic framework to represent the text. However, the semantic framework is in a two-dimensional table form, which is not flexible and extensible enough to represent potential safety hazards such as power equipment. At the same time, the definition of semantic framework relies heavily on expert experience, and it is difficult to explore the inherent complexity, relevance and regularity of hidden danger records. Reference [9] analyzes transformer faults in substation through text mining, and evaluates how various factors cause tripping problems. In order to solve the limitation of traditional expert experience, some researches use machine learning algorithm to mine data, and automatically mine the rules of keywords in text records by artificial intelligence method, and use statistical features to express the text vectorially.

This paper presents a novel approach to manage and analyze text information related to hidden dangers, particularly in the context of substations. Instead of traditional two-dimensional semantic frameworks, it employs a knowledge map's relational graph structure to represent text information and its interconnections. To efficiently store and index hidden danger data, an elastic search engine is developed, taking into account the inherent logic of substation-related hidden danger text information. A unique method, the Hidden Markov Model-HMM-VA (Hidden Markov Model-Viterbi Algorithm), is introduced for automatically extracting information necessary for constructing a knowledge map from a hidden danger corpus. Utilizing a secondary graph database and Echart rendering technology, the system dynamically generates knowledge maps and conducts correlation analysis of hidden danger records. A case study involving substation power hidden danger data in a specific region demonstrates that this method is effective in managing hidden danger texts and mitigating potential power safety risks.

**1 Substation security risks data extraction and storage**

Generally, the hidden dangers of substations in power system are recorded by manual entry into the hidden danger investigation and management table, including multi-dimensional and massive text data such as hidden danger investigation time, hidden danger equipment information, operation and maintenance management violation information, prevention and control measures, etc. In order to dynamically analyze hidden dangers by using these unstructured data, firstly, the information of hidden dangers is extracted, and the unstructured data is converted into JavaScript object notation format, and then an elastic search engine is designed to realize efficient data storage.

1.1 Substation equipment hidden danger information extraction

Substation operation involves many departments and units, as well as many kinds of facilities, equipment and circuits, which is a specialized operation within the power system with complicated hidden dangers[13]. Although there are relevant electric power specifications that summarize the components and corresponding phenomena that may cause hidden problems in the form of tables, the hidden problems are so complex and diverse that it is difficult to summarize them comprehensively in the tables in the specifications.

Substation general hidden trouble investigation and management table is a special unstructured document, which has the characteristics of semi-structured document in form, but the data flow is actually unstructured. Except the information in the document, all the original table lines are replaced by spaces and line breaks, and the real data is mixed with these spaces and line breaks, which increases the difficulty of computer processing. From the data category, the data in unstructured form documents can be divided into header area and data area. The header area indicates the nature and category of data, and the data area indicates the actual value of data. For example, "detailed classification of hidden dangers" is the header area, and "switch breaker equipment" is the data area. Hidden danger data extraction is to extract all the title areas and data areas in the table, and data organization is to establish the semantic relationship between the title areas and data areas and the semantic relationship between related title areas, and store them in JSON format.

The data extraction process is shown below: Firstly, the data extraction step is carried out, that is, the unstructured data is extracted by a standardized JSON generator to form a JSON file, and the corresponding data entities are extracted, such as hidden danger investigation time, investigation place, equipment name, accident hidden danger content, violation information, evaluation level, prevention and control measures, etc. Then, the extracted JSON file is read and parsed by using the guide tool, and the classification and attributes are divided according to the parsed features, and each entity is classified into corresponding categories and given attributes.

### 1.2 hidden danger data storage based on Elastic Search engine

Due to the dense substation equipment, hidden danger data increases exponentially with the increase of equipment, and the automatic generation and association of knowledge maps require efficient data storage and retrieval operations, so this paper designs a real-time elastic distributed search engine based on Elastic Search. We show below the data storage flow of potential safety hazards. Because the search process needs a lot of iterative calculation, in order to improve the performance, an inverted index method is constructed on the service host machine, which is composed of all non-repetitive words in JSON, and the mapping between words and the document list containing them is established. Each record in the document list includes the document identification number ID(identity card), the frequency of occurrence, the position where the word appears in the document, and so on.

Build multiple index slices, each slice is stored with independent hidden danger data, and each slice has only one data [14-15]. Through the routing formula, the serial number of the fragment is calculated by using the hash modulo method, and then the fragment information to which the fragment belongs can be queried through the index metadata of the master node, and the information of its Internet protocol can be continuously obtained, and then the information of JSON is forwarded to each slave node for storage.

An example of inverted full-text indexing of knowledge map is shown in Figure 1. Existing similar data include information such as abnormality of main transformer, oil leakage of main transformer, capacity of main transformer and shutdown of main transformer. Inverted indexing is to split the words in the data, build a table, and then disassemble the keywords, such as "key value" to index "key". When the "main transformer" is entered, it will be split into two keywords: "main" and "transformer", which will be used to retrieve data in the inverted index table and return the results.

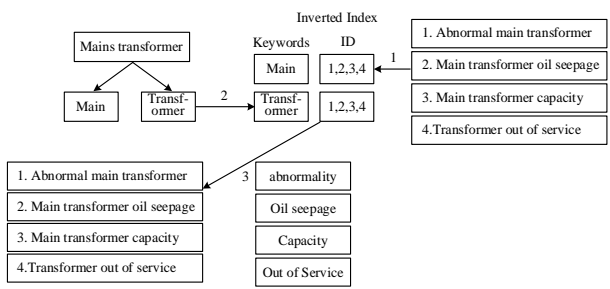

Fig.1 Example of inverted full-text index of hidden danger knowledge

Using high-performance Lucene information search library to handle chip-level index queries and maintain related index files, Elastic Search writes human function metadata on Lucene, such as mapping of hidden danger fields, index configuration and other cluster metadata. Several segments and submission points constitute Lucene index, and each segment is an inverted index. The submission point is used to record the available segments, and all available segments can be obtained through the submission point and queries can be made on the segments.

## 2 HMM-VA-based text segmentation model for security risks

Each Chinese character in power equipment information has its own word-formation position. Word-formation positions can be represented by four kinds of labels, namely, $B$ stands for the first word, $M$ stands for the middle word, $E$ stands for the last word, and $S$ stands for single word formation. Each piece of information in the equipment information constitutes an observation sequence, and the word formation of each word constitutes a state sequence. Word segmentation of equipment information can be transformed into word-formation tagging. Based on the processed corpus, the parameter information $\lambda = (\pi, X, Y)$ of hidden Markov model is obtained, and then the word-formation tagging sequence of the text to be segmented is obtained by Viterbi algorithm (VA).

The parameters of hidden Markov model include observation sequence $O = \{o_1, o_2, \cdots, o_t\}$; State sequence $Q = \{q_1, q_2, \cdots, q_t\}$; The initial state probability set $\pi$ represents the probability of each state of the model at the initial moment; The state transition probability matrix $X$ represents the probability that the model transitions between states; The observation probability matrix $Y$ represents the probability that the model obtains each observation value according to the current state.

The initial state probability set $\pi$ can be expressed as

$$\pi = \{\pi_i = P(q_i = o_i), 1 \leqslant i \leqslant N\} \quad (1)$$

Where: $i$ is the $i$-th observation state; $N$ is the maximum number of observation states.

The state transition probability matrix $X$ can be expressed as

$$X = \begin{bmatrix} 0 & P(M/B) & P(E/B) & 0 \\ 0 & P(M/M) & P(E/M) & 0 \\ P(B/E) & 0 & 0 & P(S/E) \\ P(B/S) & 0 & 0 & P(S/S) \end{bmatrix} \quad (2)$$

Where $z$ is the word position sequence of a word, and $z = (B, M, E, S)$.

The observation probability matrix $Y$ can be expressed as

$$Y = \begin{bmatrix} P(o_1/B) & P(o_2/B) & \cdots & P(o_n/B) \\ P(o_1/M) & P(o_2/M) & \cdots & P(o_n/M) \\ P(o_1/E) & P(o_2/E) & \cdots & P(o_n/E) \\ P(o_1/S) & P(o_2/S) & \cdots & P(o_n/S) \end{bmatrix} \quad (3)$$

Where: $P(o_n/z)$ is the probability of the observed value; $o_j$ is the observed value.

After the training of hidden Markov word segmentation model, the word formation position of Chinese characters in equipment information can be predicted by VA to achieve the purpose of word segmentation. VA uses the idea of dynamic programming to solve the maximum probability hidden state sequence of a given observation sequence. The basic flow of the algorithm is as follows.

Step 1: initialization, i.e.

$$S_1(i) = \pi_i b_i(o_1) \quad (4)$$

$$\varphi_1(i) = 0 \quad (5)$$

Where: $S_1(i)$ is the maximum probability of the initial time of the observation sequence in state $i$, $1 \leqslant i \leqslant N$; $\varphi_1(i)$ is the single path with the greatest probability in state $i$ at the initial moment; The probability that $\pi_i$ is the initial state $i$; Probability that $b_i(o)$ is the observed value $o_1$.

Step 2: state transition, i.e.

$$\begin{cases} S_t(i) = \max_{1 \leq j \leq N}[S_{t-1}(j)a_{ij}]b_i(O_t) \\ \varphi_t(i) = \arg \max_{1 \leq j \leq N}[S_{t-1}(j)a_{ij}] \end{cases} \quad (6)$$

Where: $S_t(i)$ is the maximum probability of the observation sequence $t$ in state $i$ at moment, $1 \leq i \leq N$; $\varphi_t(i)$ is the single path with the greatest probability in state $i$ at time $t$; $O_t$ is the observation sequence at time $t$, $2 \leq t \leq T$ and $T$ is the maximum time.

Step 3: outputs the maximum probability state sequence, namely

$$i_t^* = \varphi_{t+1}(i_{t+1}^*) \quad (7)$$

$$I^* = (i_1^*, i_2^*, \cdots, i_T^*) \quad (8)$$

Where: $i_t^*$ is the shortest path; $I^*$ is the last path sequence.

## 3 Dynamic analysis of security risks based on graph search

### 3.1 Construction of knowledge graph of substation safety hazards

Inspired by previous work [16-18], we use temporal knowledge graphs to store the extracted data. The knowledge graph detailing safety hazards in substations encompasses a wide range of entities and relationships. As equipment and facilities are upgraded, it's vital to regularly update this knowledge graph to maintain the precision and utility of equipment data retrieval. Utilizing graph databases for visualization enhances the readability of substation equipment information. This makes it easier for operations and maintenance managers to swiftly access essential parameters and operational data of substation equipment. Here's an outline of the process for constructing this safety hazard knowledge graph in substations. Step 1 Word segmentation and entity attribute extraction. Step 2 Relationship extraction. Step 3 Graph generation.

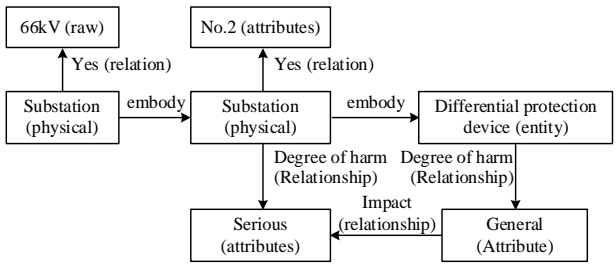

Fig.2 Relationships of entity-entity, entity-attribute, and attribute-attribute

### 3.2 Dynamic analysis of hidden dangers based on knowledge map

By integrating the Vue.js gallery into the application program, the hidden danger data stored in the Secondary graphic database is pushed to the Web for display, and the chart visualization of the data is realized by using Echarts, and the causes, categories and hazards of specific hidden dangers are visually presented to the operation and maintenance managers, so as to guide the adoption of temporary control measures and prevention and control measures.

According to the above steps, by recording the information of potential safety hazards and adding specific key information fields, such as "66kV", "Substation" and "Rainy Day", the knowledge map of potential safety hazards of 66 kV substations in this area can be dynamically generated in a personalized way. By using the correlation search of the map, the categories of potential hazards, causes of potential hazards, hazards after untreated results, possible treatment methods, violation of regulations and rules, prevention and control measures and so on can be analyzed.

## 4 Experimental example analysis

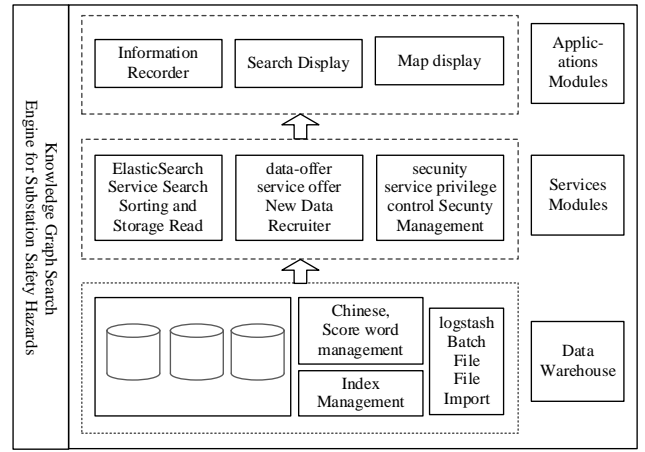

Fig.3 Search engine architecture of the hidden danger knowledge graph

The proposed HMM-VA word segmentation model is used to segment the text of potential safety hazards, and the effectiveness of the proposed method is verified by the following four examples, including the comparison of entity word segmentation methods of potential safety hazards in substations, the comparison of information retrieval performance of search engines, the knowledge map analysis of the causes of potential safety hazards in substations, and the statistical analysis and prediction of potential risks.

### 4.1 Comparison of Entity Segmentation Methods for Substation Security Risks

We compare various named entity word segmentation models, including the Boyer-Moore BM match model, the N-gram segmentation model, the Jieba model, and the newly proposed HMM-VA model. All models were tested on the same training and test sets. The results, presented in Table 1, show that the HMM-VA model, designed for power security hidden danger segmentation, surpasses the other models in terms of accuracy, recall, and F-value.

Tab.1 Comparison of effects among named entity recognition models

| Experimental model | Precision/% | Recall/% | F/% |
|---|---|---|---|
| BM matching model | 70.25 | 57.64 | 62.86 |
| N-gram seg model | 71.57 | 63.84 | 69.21 |
| Jieba model | 74.56 | 66.12 | 76.25 |
| HMM-VA | 80.25 | 75.12 | 82.54 |

### 4.2 search engine information retrieval performance comparison

The system's index performance is evaluated by comparing the indexing speed of a stand-alone index and a distributed index. The stand-alone index refers to a search engine that uses Elasticsearch's default configuration. In contrast, the distributed index is a specially designed and configured search engine, operating across 12 nodes for optimal performance within this system. The outcomes of these index performance tests are presented in Table 2, while Table 3 displays the results for search time.

As shown in Table 2, the indexing efficiency indexes of the elastic distributed search engine designed and configured in this paper are significantly better than those of the stand-alone search engine in terms of average data rate, CPU occupancy, memory occupancy, read/write rate, load rate, etc., which indicates that the proposed method effectively improves the real-time indexing of hidden danger data and meets the requirement of fast disposal efficiency in the actual substation hidden danger investigation work.

As shown in Table 3, for the search of four test keywords, the average response time of the stand-alone machine is 1,305ms, and the average response time of the search proposed in this paper is 109.5 ms, indicating that the response time of this engine is significantly lower than the original stand-alone response time, which fully proves the advantages of the search engine in the paper, and with the increase in the volume of data, the search engine has a very significant advantage over the processing of hidden safety data.

Tab.2 Index test results

| Engine | Average speed / (thousand items·s-1) | CPU occupancy rate/% | Memory occupancy/% | Read/Write Rate /% | Load Rate/% |
|---|---|---|---|---|---|
| Standalone Engine | 4 | 53 | 60.9 | 88 | 45.9 |
| This engine | 36 | 12 | 10.7 | 42 | 14.5 |

Tab.3 Search time test results

| Test keywords | Stand-alone response time/ ms | searchSystem search response time/ ms |
|---|---|---|
| Substation | 856 | 96 |
| Rainy day | 1 536 | 152 |
| Transformer | 1 345 | 86 |
| Protection regulations | 1 483 | 104 |

4.3 Knowledge mapping analysis of the causes of substation safety hazards

The information gathered about 220kV substations has led to the creation of a knowledge map, as illustrated in Figure 8, which highlights the various safety risks present in such substations. This map clearly identifies three primary categories of hazards in 220kV substations: risks to personal safety, dangers associated with equipment and facilities, and threats to electrical power safety.

(1) Personal safety hazards. For example, if the sulfur hexafluoride gas tank in the substation is not stored in the special warehouse, there is a personal safety hazard, so it is necessary to strengthen the inspection of the substation. Before it is stored in the special warehouse, no personnel are allowed to enter the warehouse, and the site environment is tested to see if there is any leakage; In the 220kV substation, there are cracks on the drainage manhole cover of the equipment site due to the crushing of vehicles during construction, and the cracks are serious, which do not conform to the "Substation Operation Regulations of State Grid Corporation", so it is necessary to replace the drainage manhole cover in time; There is no standard road fence in the construction area, which has potential safety hazards.

(2) Hidden dangers of equipment and facilities. For example, it is necessary to contact the manufacturer to replace worn-out equipment and facilities; The operation risk caused by network equipment outage is low, so the equipment and facilities of 110kV voltage level in 330kV substation need to be equipped with independent protection equipment; On-site staff should wear cotton long-sleeved overalls; There is a problem with the CPU board; Hidden dangers of equipment and facilities affect the protection function of the system.

(3) The hidden danger of power safety. For example, the power line of the 220kV Mouyin Station of the Fourth Railway violates the bolt loosening regulations, and the transformer may be grounded, which can not quickly cause the switch to trip, which may easily cause personal injury; The bolt of the ground wire clamp of Tower 8 is missing, which violates the provisions of "Line Body: Bolt Looseness" in the main table of "Operation Rules for Overhead Transmission Lines"; The oil level indicator of phase B current transformer of 220kV Dewu line in Wukeshu 220kV substation drops due to oil leakage. If it is not handled in time, it will lead to insulation breakdown of current transformer, protection action of 220kV bus differential, protection action of 220kV Dewu line, unplanned shutdown of 220kV Dewu line, No.2 main transformer and 220kV north bus in an instant, and reduce power supply load.

4.4 Statistical analysis and prediction of hidden risks

According to the statistics of substation hidden danger knowledge maps from March to July, as shown in Figure 9, there are six kinds of substation hidden dangers, such as winding deformation $E_1$, fault shutdown $E_2$, protection misoperation $E_3$, drainage line falling off $E_4$, pollution flashover and rain flashover accident $E_5$, and mechanism pressure relief $E_6$. According to the knowledge map of substation, the occurrence times of these six kinds of hidden dangers can be counted, and the occurrence rules of different hidden dangers in different months can be obtained. The hidden dangers of substation can be prevented in a targeted manner, and different measures and countermeasures can be taken to ensure the safe and stable operation of substation.

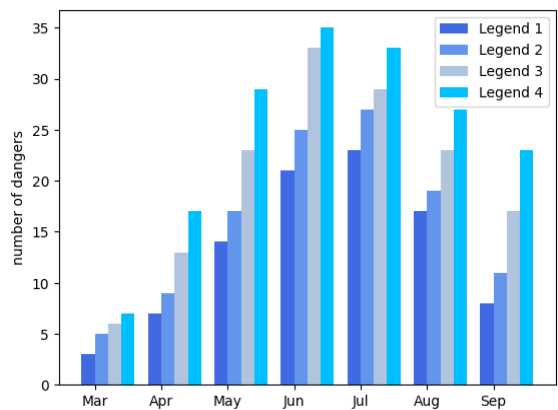

Fig.4 Statistical analysis of the number of hidden dangers in substation

The key performance characteristics of hidden dangers are described through the hidden danger map. According to the seasonal and periodic laws, the hidden dangers of a certain type of equipment or operation in the future, such as "winding deformation", are obvious in March and June, and extra attention should be paid to prevention, as well as the analysis of the other five hidden dangers. Through the statistical analysis of hidden dangers in substations, the preventive measures are mainly based on the expert experience of substations, the guidance and suggestions of "Substation Operation Regulations of State Grid Corporation of China" and "White Paper on Hidden Danger Risk Management of State Grid Corporation of China", which can predict the location of hidden dangers. For example: (1) The transformer that suffers from short circuit at 10kV line exit and short circuit impact in the near area for many times cannot effectively control the impact degree and is prone to winding deformation; (2) Transformer equipment and electromagnetic voltage transformers that run for more than 15 years are prone to malfunction and shutdown; (3) Countermeasures such as imperfect rainproof measures for transformer gas relay and pressure release valve are not implemented, which may easily lead to mis-operation of protection; (4) Looseness of clamp and crimping tube of drainage line of some equipment gradually appears, which is easy to cause drainage line to fall off and cause accidents; (5) The climbing distance of external insulation of some substation equipment does not meet the standard requirements, and it is located in a heavily polluted area, and pollution flashover and rain flashover accidents are prone to occur in rainy, snowy and foggy days; (6) The 220 kV switch hydraulic mechanism needs to be overhauled periodically, which may lead to the risk of pressure relief and forced shutdown if the overhaul period is exceeded or the construction technology is not in place.

## 5 Conclusion

In this paper, knowledge map technology and elastic distributed search engine technology are introduced, and the dynamic analysis method of substation security risks is put forward. The data and distributed storage of security risks are analyzed in detail, and the complete dynamic analysis process of hidden dangers such as word segmentation and knowledge map construction based on HMM-VA is given. The effectiveness of a knowledge map search engine in visual representation, quick information retrieval, and analytical correlation has been demonstrated through practical experiments. This approach offers valuable insights and direction for managing and mitigating security risks, showcasing its strong practical application and potential for widespread use. Future research will focus on extracting additional features from the corpus during the relationship extraction phase. This enhancement aims to refine the construction of knowledge maps, thereby elevating the precision of safety hazard analysis in substations.